\documentclass[sigconf,natbib=false]{acmart}
\usepackage{multirow}
\usepackage{subcaption}
\usepackage{graphicx}
\usepackage{caption}
\AtBeginDocument{%
  }

\setcopyright{acmlicensed}
\copyrightyear{2026}
\acmYear{2026}
\setcopyright{cc}
\setcctype{by}
\acmConference[SAC '26]{The 41st ACM/SIGAPP Symposium on Applied Computing}{March 23--27, 2026}{Thessaloniki, Greece}
\acmBooktitle{The 41st ACM/SIGAPP Symposium on Applied Computing (SAC '26), March 23--27, 2026, Thessaloniki, Greece}
\acmPrice{}
\acmDOI{10.1145/3748522.3779754}
\acmISBN{979-8-4007-2294-3/2026/03}

\RequirePackage[
  datamodel=acmdatamodel,
  style=acmnumeric,
  ]{biblatex}

\begin{document}

%%
%% The "title" command has an optional parameter,
%% allowing the author to define a "short title" to be used in page headers.
\title{Optimizing the Training Diet: \\ Data Mixture Search for Robust Time Series Forecasting}
% Please make sure that the short title does not exceed the width of one column
\renewcommand{\shorttitle}{Optimizing the Training Diet}

%%
%% The "author" command and its associated commands are used to define
%% the authors and their affiliations.
%% Of note is the shared affiliation of the first two authors, and the
%% "authornote" and "authornotemark" commands
%% used to denote shared contribution to the research.
\author{Federico Pennino}
\email{federico.pennino2@unibo.it}
\orcid{1234-5678-9012}
\affiliation{%
  \institution{DISI - Università di Bologna}
  \city{Bologna}
  \state{Emilia-Romagna}
  \country{Italy}
}

\author{Maurizio Gabbrielli}
\email{maurizio.gabbrielli@unibo.it}
\affiliation{%
  \institution{DISI - Università di Bologna}
  \city{Bologna}
  \state{Emilia-Romagna}
  \country{Italy}
}

%This command displays author info in page headers
% Please use the following convention:
% One author: J. Smith
% Two authors: J. Smith and I. Jones
% Three and more authors: J. Smith et al.
\renewcommand{\shortauthors}{F. Pennino and M. Gabbrielli}

%%
%% The abstract is a short summary of the work to be presented in the
%% article.
\begin{abstract}
The standard paradigm for training deep learning models on sensor data assumes that more data is always better. However, raw sensor streams are often imbalanced and contain significant redundancy, meaning that not all data points contribute equally to model generalization. In this paper, we show that, in some cases, “less is more” when considering datasets. We do this by reframing the data selection problem: rather than tuning model hyperparameters, we fix the model and optimize the composition of the training data itself. We introduce a framework for discovering the optimal “training diet” from a large, unlabeled time series corpus.

Our framework first uses a large-scale encoder and k-means clustering to partition the dataset into distinct, behaviorally consistent clusters. These clusters represent the fundamental ‘ingredients’ available for training. We then employ the Optuna optimization framework to search the high-dimensional space of possible data mixtures. For each trial, Optuna proposes a specific sampling ratio for each cluster, and a new training set is constructed based on this recipe. A smaller target model is then trained and evaluated. Our experiments reveal that this data-centric search consistently discovers data mixtures that yield models with significantly higher performance compared to baselines trained on the entire dataset. Specifically --- evaluated on PMSM dataset --- our method improved performance from a baseline $MSE$ of 1.70 to 1.37, a 19.41\% improvement.
\end{abstract}

%%
%% The code below is generated by the tool at http://dl.acm.org/ccs.cfm.
%% Please copy and paste the code instead of the example below.
%%


%%
%% Keywords. The author(s) should pick words that accurately describe
%% the work being presented. Separate the keywords with commas.

%%
%% This command processes the author and affiliation and title
%% information and builds the first part of the formatted document.
\maketitle

\section{Introduction}
In many machine learning applications, including time series modeling, the prevailing assumption is that larger datasets naturally translate into better model performance. Yet, real-world sensor streams often exhibit severe imbalances, highly redundant patterns, and complex latent regimes that contribute unequally to the model’s generalization capacity \cite{10982066}. Simply increasing the dataset size may amplify redundancy, leading to inefficient training. Moreover, without careful curation, the additional data can obscure informative patterns, overwhelm optimization procedures, and hinder the model’s ability to learn rare but critical behaviors.

Recent advances in data-centric learning \cite{2024arXiv240416886S} have challenged this model-centric paradigm by focusing on the composition of the training data. In the context of large language models, methods such as CLIMB \cite{2025arXiv250413161D} (Clustering-based Iterative Data Mixture Bootstrapping) have demonstrated that actively selecting and mixing data clusters can substantially improve pre-training efficiency and downstream task performance. CLIMB leverages an embedding-based clustering of massive text corpora and performs an iterative search of mixture weights, guided by proxy models and learned performance predictors.

However, extending such strategies to time series domains is non-trivial. While textual data also exhibits temporal dependencies and high dimensionality, time series present unique challenges: they are inherently continuous, often multivariate with complex inter-variable dependencies, and contain latent regimes that do not decompose cleanly into discrete semantic categories as words or tokens do in language. Moreover, temporal coherence and channel alignment introduce challenges for representation learning and clustering.

In this work, we propose a novel direct data-centric optimization framework for time series learning, inspired by—but fundamentally distinct from—prior work in language model pre-training. We employ \texttt{MOMENT} \cite{goswami2024moment}, a large-scale foundational encoder \cite{2024arXiv240103717T, 2024arXiv240113912M}, to transform raw time series into dense, task-agnostic embeddings. These embeddings naturally capture behavioral similarities, allowing us to partition the dataset into meaningful clusters via K-Means clustering \cite{wu2012advances}. The resulting clusters serve as atomic units for data selection, and we pose the data composition problem as a hyperparameter optimization task. Using Optuna \cite{2019arXiv190710902A}, we search for the optimal sample allocation across clusters that maximizes the target model’s performance on the downstream task.

\begin{figure*}[t]
    \centering
    \includegraphics[width=0.8\linewidth]{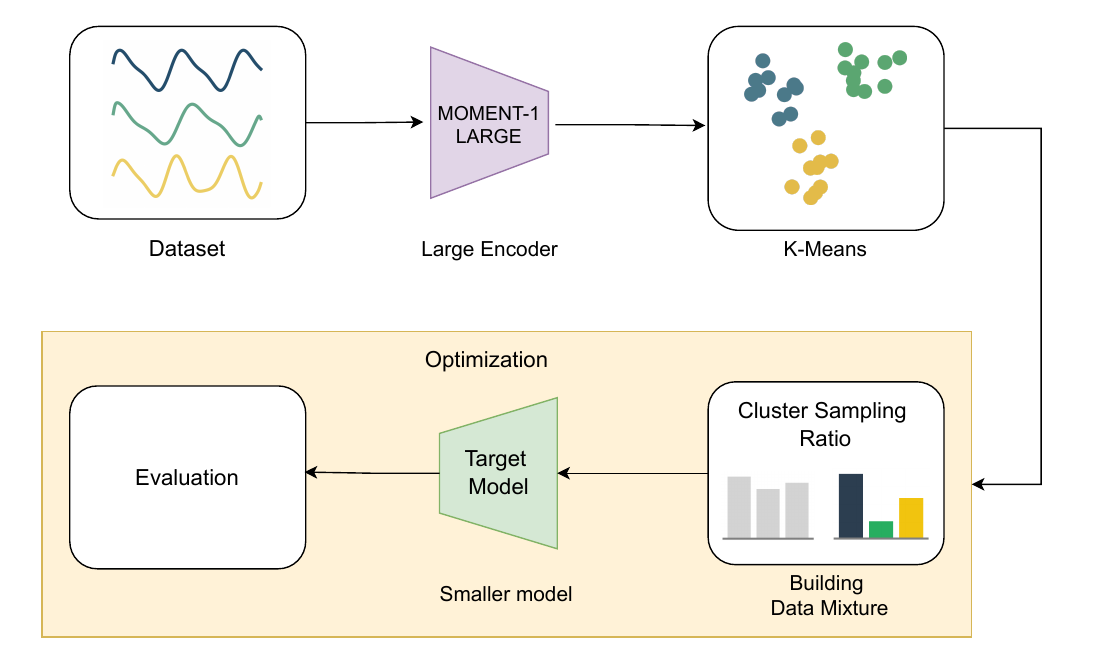}
    \caption{Pipeline of the training data mixture optimization. The full time-series dataset is embedded using a large pre-trained encoder and clustered using k-means into consistent groups. Optimal cluster sampling ratios are determined using Optuna’s optimization, generating multiple candidate training sets. For each mixture, a fixed-target model is trained and evaluated to guide the search toward data compositions that improve model generalization.}
    \label{fig:pipeline}
\end{figure*}

Unlike CLIMB’s iterative bootstrapping, which uses proxy models and predictive regressors, our approach operates in a fully supervised loop, directly evaluating each candidate data mixture against the final model. This yields a data mixture that is directly tuned to the target task and model, without additional surrogate models or multi-stage heuristics. In this sense, our method remains both computationally tractable and easily adaptable across diverse time-series learning scenarios.

When applied to the PMSM dataset \cite{wilhelm_kirchg_ssner_oliver_wallscheid_joachim_b_cker_2021}, our framework consistently identifies data mixtures that outperform the baseline trained on the full dataset, improving performance by 19.41\% while using just 42.6\% of the original data. More broadly, this work demonstrates how a carefully structured encode-cluster-optimize pipeline enables systematic data selection for time series. In this area, existing literature remains largely underexplored compared to text and image domains.

In summary, our main contributions are as follows:

\begin{itemize}
    \item We introduce a data-centric training pipeline --- illustrated in \figurename~\ref{fig:pipeline} --- for time series that embeds raw sensor streams with a large-scale encoder, discovers latent behavioral regimes via K-Means clustering, and directly optimizes the sampling composition.
    \item We formulate the selection of cluster sampling ratios as a hyperparameter optimization problem and efficiently solve it with Optuna, achieving consistent, significant gains over conventional full-data training.
    \item We employ a quantitative evaluation with an LLM-as-a-reviewer step, where a large multimodal model provides qualitative human-interpretable validation of our sampling strategy.
\end{itemize}

We deliberately focus our analysis on the PMSM dataset to conduct a thorough, controlled investigation of our proposed framework. This single-dataset approach allows us to perform comprehensive cluster analysis. By establishing the effectiveness of our encode-cluster-optimize pipeline on a well-characterized benchmark, we provide a solid foundation for future multi-dataset validation studies. The findings of this work challenge the prevailing data-maximization paradigm in time series modeling and demonstrate that intelligent data selection --- rather than sheer data volume --- can unlock superior model performance and generalization.

\section{Related Work}
The composition and quality of the training dataset are universally acknowledged as critical factors influencing the performance and generalization capabilities of machine learning models. While random sampling has been a long-standing default, there is a growing consensus that more sophisticated data selection and scheduling strategies can yield significant improvements.  

Recent research increasingly uses clustering of data points (in an embedding or feature space) to inform data sampling strategies or curricula during training. One prominent example in NLP is the ClusterClip method for large language models \cite{2024arXiv240214526S}. ClusterClip clusters the training corpus by semantic similarity (using pre-trained text embeddings) to approximate the data distribution, then adjusts the sampling of mini-batches to be initially uniform across clusters. 

A conceptually related approach in LLM pre-training is CLIMB \cite{2025arXiv250413161D}, which focuses on optimizing the mixture of data from different sources or clusters. While CLIMB's primary goal is to determine the ideal blend of data for pre-training (e.g., using token counts from different data clusters), its core principle of cluster-driven data composition is highly relevant. In their framework, a separate "predictor" model is trained to map data mixture compositions (as input features) to the final LLM performance metrics (as target labels). Our proposed method shares this fundamental goal of finding an optimal data blend. However, instead of training an explicit predictor, our Optuna-based framework implicitly defines such a relationship through its objective function, directly searching for the optimal number of time-series samples from each cluster to maximize performance on a specific downstream task.

In other machine learning fields, such as computer vision, the same idea of improving the quality of the training dataset has led to the use of clustering to design curricula for generative models. In \cite{2019arXiv190611594Z}, the authors propose a clustering-based curriculum for GAN training: data points from dense, central clusters (i.e., prototypical examples) are fed to the GAN early on, whereas more peripheral (outlier) clusters are introduced gradually in later rounds. This strategy of progressing from “easy” core samples to “harder” outliers improved GAN convergence on noisy image datasets.

Despite these advances in text and image domains, time series data selection remains largely underexplored. Time series present unique challenges that distinguish them from other data modalities: they are inherently high-dimensional with complex temporal dependencies, often multivariate with intricate cross-channel relationships, and contain latent behavioral regimes that do not decompose cleanly into discrete semantic categories like text.

\cite{2019arXiv190412887K} demonstrated curriculum learning's effectiveness for financial forecasting, achieving ~30\% accuracy improvements on Microsoft's revenue data across 60 regions. Their encoder-decoder LSTM architecture with temporal complexity ordering established feasibility for medium-scale time series. However, this remains essentially the only major application of curriculum learning and specifically to time series mixture optimization, representing a critical research gap. 
Instead, despite extensive development in computer vision and natural language processing, core set selection for time series data has yet to be explored in the literature. This work fills this gap by presenting a pipeline that allows for systematic selection of the best dataset for a certain task.

This is the reason our approach differs from existing methods in several key ways: (1) Rather than CLIMB's iterative predictor-based approach, we directly optimize cluster sampling ratios through the target model's performance; (2) We utilize a pre-trained foundational encoder specifically designed for time series, ensuring that clusters capture relevant temporal dynamics rather than superficial similarities; (3) Our framework discovers data mixtures explicitly tailored to the downstream forecasting task, rather than general-purpose data balancing strategies.

\section{Methodology}
\subsection{Pipeline overview}
Our data-centric optimization framework operates on the premise that not all training samples contribute equally to the forecasting model performance. Rather than accepting the original data distribution, we systematically discover which behavioral patterns within the dataset are most valuable for the target forecasting task. The framework consists of three main stages: embedding, clustering, and optimization. 

As illustrated in \figurename~\ref{fig:pipeline}, we first transform the heterogeneous time series data into a unified representation space where behavioral similarities can be quantified. We then partition this space into distinct operational regimes, which serve as the atomic units for data selection. Finally, we search for the optimal composition of these regimes that maximizes downstream model performance.

The embedding process begins by using a large pre-trained encoder, \texttt{MOMENT-1 large} \cite{goswami2024moment}, to transform the entire dataset into dense, task-agnostic embeddings. This foundational model captures temporal dynamics and cross-channel dependencies, creating representations where time series with similar behavioral patterns cluster naturally in the embedding space. Subsequently, these embeddings are partitioned into distinct clusters using K-Means, with each cluster representing a coherent operational regime or behavioral pattern within the original data.

The optimization process, managed by Optuna \cite{2019arXiv190710902A}, is formulated as a search for the optimal set of sampling weights, $\{w_1,w_2,...,w_k\}$, corresponding to each of the K clusters derived from the dataset. Each weight, $w_k$, is a continuous hyperparameter within the search space $[0, 1]$.

For each trial within the optimization loop, Optuna proposes a vector of weights. A new training data mixture is then constructed by sampling $n_k$ elements from each cluster $C_k$. The number of elements to be sampled from a given cluster is calculated as the product of the cluster's total size and its assigned weight, formally expressed as shown in \equationautorefname~\ref{eq:weight}.

\begin{equation} \label{eq:weight}
    n_k = C_k \cdot w_k
\end{equation}

The resulting data mixture is then used to train the target model, providing a performance score that guides Optuna's subsequent search for an improved weight configuration.
\subsection{Time-series Embedding}

Time series embedding is a technique for converting raw, variable-length time series data into fixed-length numerical vectors \cite{2023arXiv231004486F, 2025arXiv250113392I}. The purpose is to capture key temporal dynamics and patterns in a compact format, making subsequent tasks like clustering more effective. This conversion is typically done using neural network encoders, such as the MOMENT \cite{2019arXiv190710902A} model used in this work, which are trained to extract relevant information into these vectors.

In our framework, embedding is a necessary first step. It organizes the raw data into a structured space where similarities between time series are represented by proximity. This enables us to utilize K-Means to categorize the data into meaningful clusters. These clusters then serve as the basic units for our data selection process, allowing us to utilize Optuna to search for the optimal data mixture for training our target model.

\begin{table*}[t]
\centering
\caption{Performance comparison across different model architectures on the PMSM thermal prediction task. Our approach uses PatchTST trained on either the full dataset or the optimized data mixture (43\% of original size). The data mixture achieves the best overall performance with a 19.41\% improvement in average MSE, demonstrating superior prediction accuracy across most engine components while using less than half the training data. }
\label{tab:mse_models}
\begin{tabular}{l|cccc|cccc|c}
\toprule
 & \multicolumn{4}{c|}{\textbf{MSE}} & \multicolumn{4}{c|}{\textbf{MAE}} & \textbf{Avg. MSE} \\
\cmidrule(lr){2-5} \cmidrule(lr){6-9}
\textbf{Model} & \textbf{Yoke} & \textbf{Tooth} & \textbf{Winding} & \textbf{PM} & \textbf{Yoke} & \textbf{Tooth} & \textbf{Winding} & \textbf{PM} & \\
\midrule
TCN   \cite{9296842}              & 6.90 & 2.84 & 1.80 & 0.65 & 5.24 & 6.24 & 7.92  & 5.84 & 3.04 \\
LSTM \cite{2022arXiv220800293L}               & 1.34 & 2.23 & 4.86 & 1.52 & 4.03 & 4.89 & 8.47  & 5.62 & 2.49 \\
BiLSTM   \cite{2022arXiv220800293L}           & 2.04 & 2.82 & 5.28 & 5.61 & 4.58 & 4.89 & 10.09 & 6.00 & 3.94 \\
AttentionLSTM  \cite{2022arXiv220800293L}     & 1.04 & 1.84 & 2.82 & \textbf{1.17} & 3.51 & 6.05 & 8.75  & 3.05 & 1.72 \\
Our (Full dataset)  & 0.84 & 1.13 & \textbf{1.31} & 3.53 & 0.68 & 0.77 & 0.89  & 1.53 & 1.70 \\
Our (Data-Mixture)  & \textbf{0.65} & \textbf{0.91} & \textbf{1.31} & 2.60 & \textbf{0.60} & \textbf{0.68} & \textbf{0.87}  & \textbf{1.32} & \textbf{1.37} \\
\bottomrule
\end{tabular}
\end{table*}

\subsection{Dataset and Preprocessing}
\label{sec:dataset}

In our study, we used the publicly available PMSM dataset \cite{wilhelm_kirchg_ssner_oliver_wallscheid_joachim_b_cker_2021}, which provides time-series data from an electric motor operating under diverse loads. The goal is to predict four key thermal metrics --- specifically the temperatures of the stator winding, tooth, yoke, and the permanent magnet --- using measurements such as d- and q-axis current and voltage, motor speed, and torque. As a preprocessing strategy, we implement the one presented in \cite{9296842}.
However, compared to the original work, we divided the entire dataset into windows of 300 timesteps to enhance the model's capability to capture long-range relationships. 

We computed Exponentially Weighted Moving Averages (EWMA) for measured and derived features using spans equal to [200, 500, 1500, 4000] to incorporate temporal context and model trends over various time scales. All features and target variables were normalized to a consistent range of zero to one using a \texttt{min-max scaler}, which was fitted only on the training data to prevent any information leakage.

The data corresponding to \texttt{profile\_id=65} was used as the test set, and the data corresponding to \texttt{profile\_id=[18, 39, 46, 56, 75]} was used for the validation set, ensuring our evaluation is performed on a completely unseen motor duty cycle. The remaining profiles were used for training. This results in a total $\simeq 1.1$ million of training windows. The evaluation set was composed of $\simeq 85.000 $ windows and the test set is composed by $\simeq 40.000$ windows.

\subsection{Experiment setup}

Our experimental setup is designed to effectively evaluate the data mixture optimization framework presented in \figurename~\ref{fig:pipeline}.

Firstly, we encode the time series into a task-agnostic embedding space using the \texttt{MOMENT-1 large} encoder. The resulting embeddings were then partitioned into 36 distinct groups using K-Means clustering, with $k=36$ chosen through manual tuning.

Optuna proposed a set of 36 cluster weights for each trial, each a floating-point number between 0.0 and 1.0. Based on these weights, a new training data mixture was sampled. A fixed-architecture target model was then trained on this data mixture. The target model is a PatchTST \cite{Yuqietal-2023-PatchTST} with sinusoidal position embeddings, configured with an embedding dimension of 256, 8 attention heads, a patch size of 30, and a convolutional layer used for patching the input time series. We decided to use this architecture to reduce the model's training throughput. 

The optimization was configured to run for 100 trials. We utilized Optuna's Tree-structured Parzen Estimator (TPE) sampler to navigate the search space and find the optimal mixture efficiently. The search was parallelized across 4 jobs to accelerate the discovery process. The study's objective was to minimize the Mean Squared Error (MSE).

Training for each trial was performed using the Adam optimizer with a learning rate of $1\cdot10^{-4}$ and a batch size of $1024$. A learning rate scheduler was employed, featuring a linear warmup for the first 30\% of the total training steps, followed by a linear decay. Transformer architecture \cite{2017arXiv170603762V} is significantly faster than LSTM \cite{10.1162/neco.1997.9.8.1735} and CNN \cite{2015arXiv151108458O}, this is the reason we adopt this architecture in this work. We fixed the training parameters for all the models.

The total number of training epochs was dynamically adjusted for each trial to ensure that the model consistently processed a total of 360 million training tokens, where a token is one time series patch passed through the model. This normalization of the training budget means that models trained on smaller, more targeted data mixtures iterate over their respective datasets more frequently. 

All experiments were conducted using PyTorch \cite{paszke2017automatic} on a cluster equipped with four NVIDIA L40s GPUs and 512 GB of RAM. The optimization process took roughly one day of computation.

\section{Discussion}
\subsection{Performance}

\begin{figure*}[t!]
    \centering
    % Panel (a)
    \begin{subfigure}[t]{0.39\textwidth}
        \centering
        \includegraphics[width=\linewidth, height=7cm]{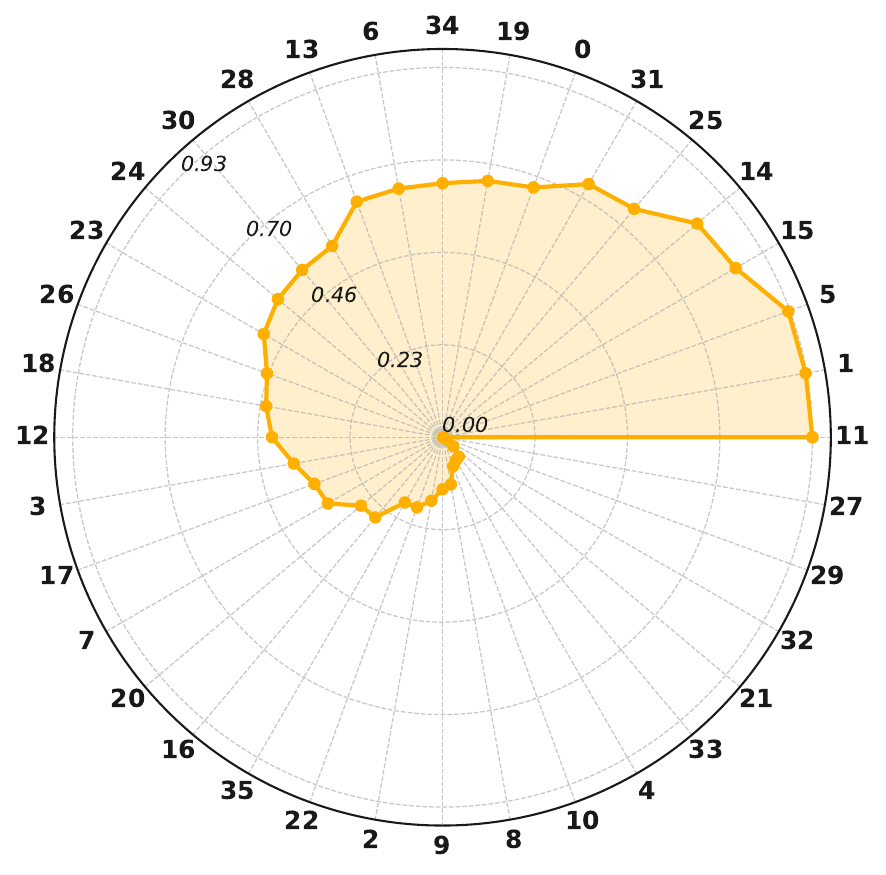}
        \caption{}
        \label{fig:radar}
    \end{subfigure}
    \hfill
    % Panel (b)
    \begin{subfigure}[t]{0.6\textwidth}
        \centering
        \includegraphics[width=\linewidth, height=6.8cm]{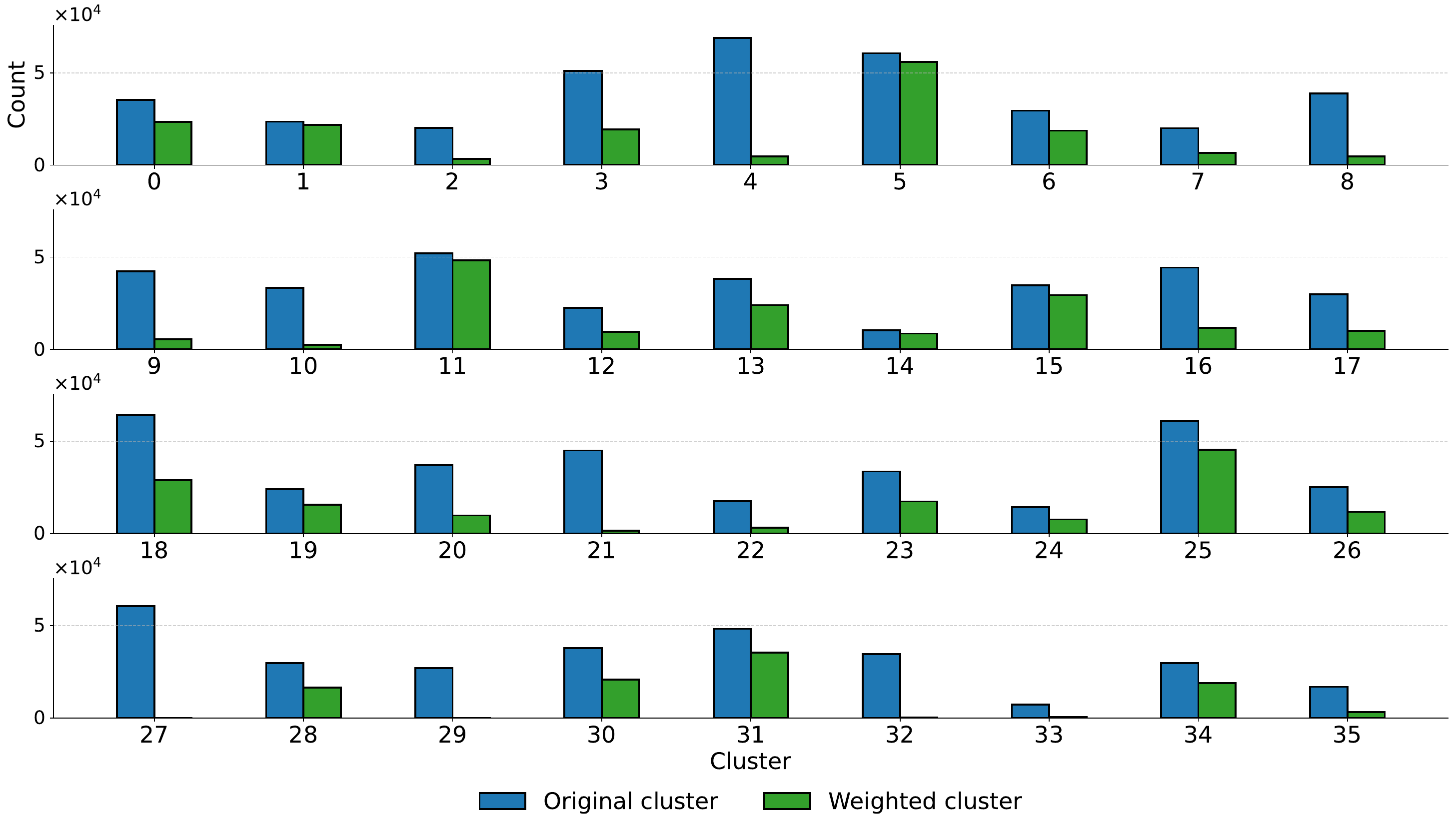}
        \caption{}
        \label{fig:bar_plot}
    \end{subfigure}

    \caption{
      \textbf{(a)} Radar chart of optimized cluster weights arranged in descending order. Each axis corresponds to a cluster, and the radial distance represents the relative weight. The chart highlights the distribution of weight values across all clusters, facilitating comparison of their relative importance.
      \textbf{(b)} Comparison of original versus weighted sample counts across the 36 clusters (0–35). 
    }
    \label{fig:combined}
\end{figure*}

\begin{figure}[b]
    \centering
    \includegraphics[width=\linewidth]{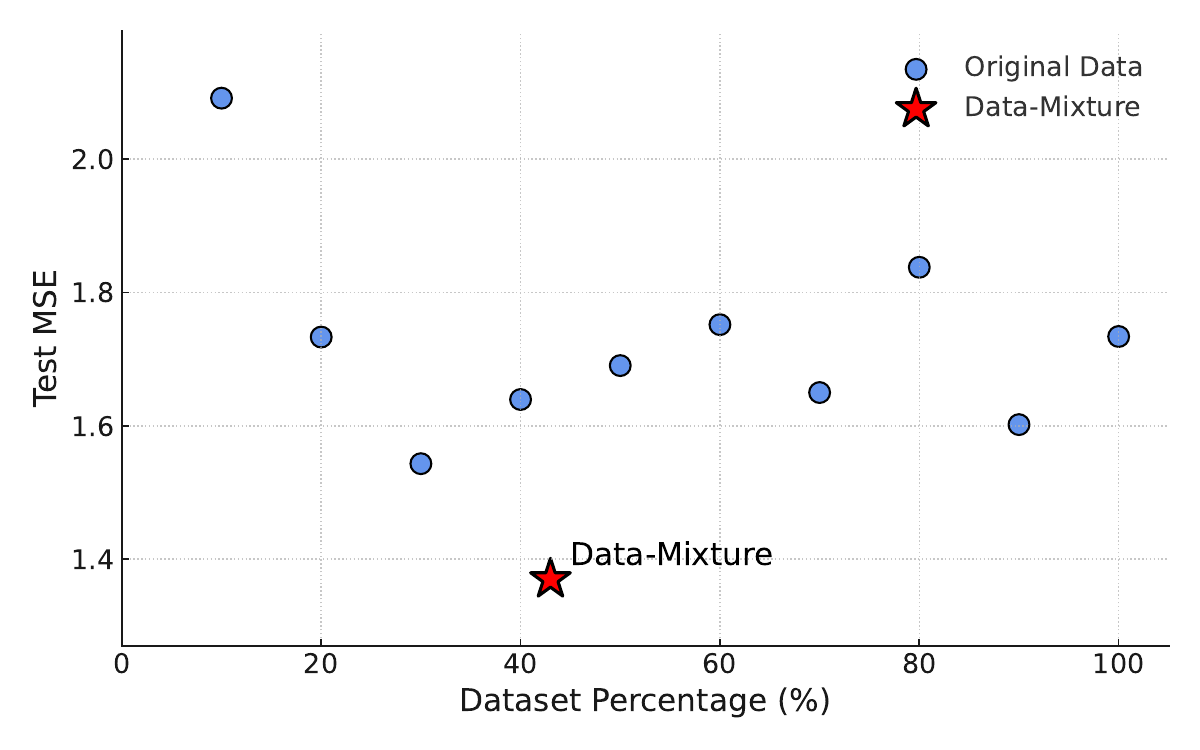}
    \caption{Test MSE as a function of training dataset size. The blue circles show the performance of the PatchTST model trained on various percentages of the original dataset, while the red star represents the model's superior performance when trained on our optimized data mixture. The data-mixture, which is only ~43\% of the size of the full dataset, achieves a significantly lower MSE (1.37) than training on 100\% of the original data (1.73).}
    \label{fig:dataset_increasing}
\end{figure}

The optimization produces a new dataset 2.3 times smaller than the original complete training set. The results, summarized in \tablename~\ref{tab:mse_models}, demonstrate the effectiveness of our proposed data-mixture optimization pipeline. We report the best value for each architecture independently from the window size. Specifically, PatchTST was trained with 300-timestep windows while traditional architectures (LSTM, BiLSTM, AttentionLSTM, TCN) used 180-timestep windows. This window-size differentiation is crucial for fair comparison, as our experiments show that extending the window size to 300 timesteps consistently degrades performance across recurrent and convolutional architectures.

Furthermore, as shown in the \figurename~\ref{fig:dataset_increasing}, our optimized data-mixture (indicated by the red star) achieves a lower MSE than the same model trained on various random percentages of the original dataset, including the full training set. This confirms that the optimized subset isn't only better than using a random subset of the same size, but also better than using any random subset.

As shown in \tablename~\ref{tab:mse_models}, our data-mixture approach significantly outperforms all the other models trained on the entire dataset. This highlights the benefit of our data selection strategy, which enables more efficient and targeted training. The model trained on the data mixture excels at predicting temperature for the Yoke and Tooth, achieving the best scores for both MSE and MAE. While the MSE for the Winding component (1.33) is on pair the model trained on the full dataset (1.31), its corresponding MAE (0.87) is the lowest among all models, indicating strong predictive accuracy. However, LSTMs perform best at predicting the temperature of Permanent Magnet (PM) components, even though our solution significantly improves upon PatchTST, which was trained on the full dataset.

\subsection{Cluster weights analysis}

To gain insight into the produced data mixture, we performed an analysis of the weight distribution. The data-centric pipeline proposed in this work produced two complementary outcomes that, together, explain the observed improvement in performance of the PatchTST target model.  First, the Optuna search discovered a non-uniform set of sampling weights (see the radar plot in \figurename~\ref{fig:radar}), effectively pruning entire clusters while strongly amplifying others.  Second, those weights reshaped the empirical training distribution into one that is markedly less skewed than the raw data (compare the paired bar charts in \figurename~\ref{fig:bar_plot}).  Taken together, these effects illuminate how a relatively small change in \emph{which} samples are shown to the model can matter as much as architectural ingenuity or longer training schedules.

The optimization process produced a highly differentiated set of weights for the 36 data clusters, as illustrated in \figurename~\ref{fig:radar}. The weights, representing the sampling proportion for each cluster, span from nearly 0.0 to almost 1.0. This variance underscores that the value of data for the given prediction task is not evenly distributed across the dataset's operational modes. Specific clusters were identified as highly valuable, receiving weights close to unity (e.g., clusters 11, 1, 5, 15), ensuring their near-complete inclusion in the training mixture. Conversely, other clusters were assigned very low weights, indicating that their contribution to the model's generalization was minimal or even detrimental.

The most striking example is Cluster 27, which received a weight approaching zero. An examination of the original data distribution (\figurename~\ref{fig:bar_plot}) reveals that Cluster 27 is one of the most populous clusters in the entire dataset. The optimization framework effectively determined that, despite its large volume, the information in this cluster was redundant or irrelevant for predicting thermal metrics on unseen validation profiles. By drastically down-sampling this cluster, the framework prioritized data quality over sheer quantity.

In general, the dataset transformation is visualized in \figurename~\ref{fig:bar_plot}. The initial dataset was heavily skewed, with a few clusters containing the vast majority of data points while many others were sparsely populated. After applying the optimal weights, the resulting training set is significantly more balanced. The initially dominant clusters (e.g., 4, 27, 18) are reduced in size to be comparable with other, more informative clusters.

\section{Interpretability}

\begin{figure}
    \centering
    \includegraphics[width=1.05\linewidth]{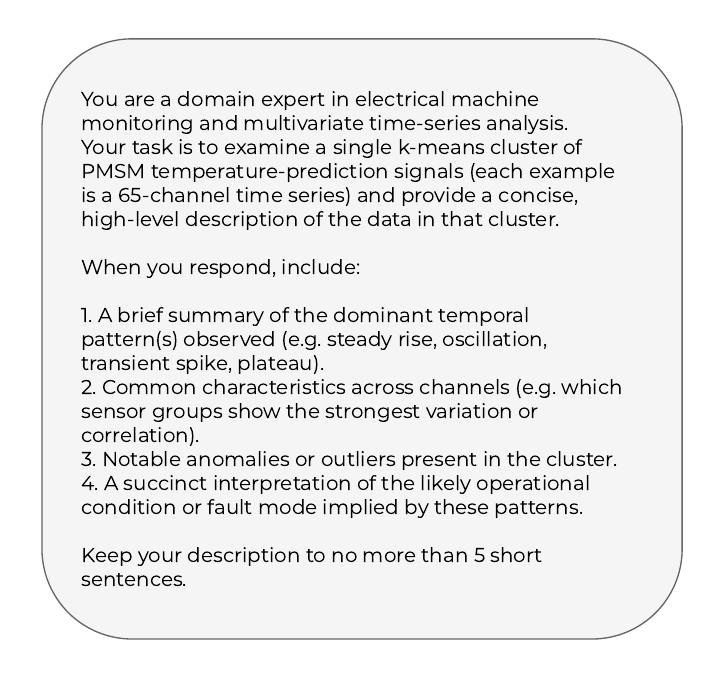}
    \caption{In this figure, we show the prompt used for the LLM-as-reviewer. The expert is required to summarize the information in the signals.}
    \label{fig:prompt}
\end{figure}

To deepen our understanding of why certain clusters drive model performance, we conducted a qualitative “LLM review” \cite{2024arXiv241115594G} of cluster-specific behavior. We used \texttt{o4-mini} \cite{openai2022chatgpt} for this analysis. The system prompt is presented in \figurename~\ref{fig:prompt}.

For each of the three most heavily weighted clusters (11, 1, and 5) and the three least weighted clusters (27, 29, and 32), we randomly sampled ten representative time-series plots. We fed them to a large language model, asking it to characterize their dynamics and potential informational value. The LLM consistently highlighted that samples from clusters 11, 1, and 5 contain rich, structured variations—pronounced startup transients, clear load-change events, and well-defined steady-state plateaus—suggesting these data segments capture the fundamental modes of system operation. 

In contrast, the LLM described samples from clusters 27, 29, and 32 as largely uninformative: predominantly flatlines, subtle sensor noise, or simple shutdown spikes with little dynamic range. By presenting these insights alongside our Optuna-derived sampling weights, we not only reinforce why clusters 11, 1, and 5 are essential ingredients in the optimal “training diet” but also offer an interpretable rationale for deprioritizing clusters 27, 29, and 32. 

This qualitative LLM-driven analysis thus bridges the gap between raw performance metrics and human intuition, demonstrating that the clusters we up-weight truly embody informative machine behaviors while those we down-weight contribute little beyond monotony.

\section{Conclusion}

In this work, we introduced a novel data‐centric framework for time‐series forecasting that directly optimizes the composition of the training set via a mixture‐search approach. 

By leveraging a pre-trained foundational encoder (MOMENT-1) to obtain dense embeddings, partitioning the resulting representation space with K-Means, and formulating the cluster sampling ratios as hyperparameters to be tuned with Optuna, we demonstrated that a carefully curated “training diet” can substantially outperform models trained on the full dataset. 

Experiments on the PMSM benchmark showed that our optimized data mixture achieves a 19.41\% reduction in average MSE compared to the same model trained on all available data, while using fewer than half as many samples. Furthermore, we analyzed the meaning of the weighted clusters.

Overall, our encode–cluster–optimize pipeline provides a scalable, model-agnostic approach to data selection in time-series learning, challenging the prevailing data-maximization paradigm and opening new avenues for efficient, performance-driven dataset design. To support reproducibility, we provide complete source code for data preprocessing, mixture optimization, model evaluation, and plot generation for LLM analysis. All materials are available at: 

\url{https://github.com/halykoss/cluster_based_training}

\section{Limitations}

Despite its promising results, our study has several limitations. First, the framework relies on a fixed, pre-trained encoder and a predetermined number of clusters; the choice of embedding model and cluster count may significantly impact the quality of the discovered mixtures and may require task-specific tuning. Second, the Optuna-based search adds a substantial computational cost, as each trial trains a complete downstream model; this overhead may be prohibitive for extensive datasets or resource-constrained settings. Third, our evaluation is restricted to the PMSM dataset and a limited set of model architectures; further validation on diverse time‐series domains (e.g., finance, healthcare) and with alternative forecasting models is necessary to assess generalizability. Fourth, while the optimized mixtures improved average performance, specific clusters were heavily pruned, potentially discarding rare but critical regimes. Adaptive strategies to preserve minimal coverage of low-frequency behaviors warrant exploration. Finally, the method currently targets supervised forecasting tasks; extending the approach to unsupervised or multi‐task scenarios remains an open challenge. Addressing these limitations will be crucial for translating data-mixture optimization into broader, real-world applications and constitutes our primary focus for future investigations.

%%
%% Print the bibliography
%%
\printbibliography

\end{document}